\def\year{2019}\relax
\documentclass[letterpaper]{article} 
\usepackage{aaai19}  
\usepackage{times}  
\usepackage{helvet}  
\usepackage{courier}  
\usepackage{url}  
\usepackage{graphicx}  

\usepackage{epsfig,epstopdf}
\usepackage{subfigure}
\usepackage{multirow}
\usepackage{booktabs}
\usepackage{latexsym,bm}
\usepackage{enumitem,balance,mathtools}
\usepackage{wrapfig}
\usepackage{algorithm}
\usepackage{algorithmic}
\usepackage{ifpdf}
\usepackage{diagbox}
\usepackage{makecell}
\usepackage{subfigure}
\usepackage{bm}
\usepackage{amssymb}

\newcommand{\bs}{\boldsymbol}

\newcommand{\bx}{\bs{x}}

\newcommand{\br}{\bs{r}}
\newcommand{\btheta}{\bs{\theta}}

\usepackage{array}
\newcolumntype{N}{@{}m{0pt}@{}}
\newcommand{\citet}[1]{{\citeauthor{#1} \shortcite{#1}}}

\begin{document}
%
\title{Deep Recurrent Survival Analysis}
\author{Kan Ren, Jiarui Qin, Lei Zheng, Zhengyu Yang, \\
	{\bf \Large Weinan Zhang, Lin Qiu, Yong Yu}  \\
	APEX Data \& Knowledge Management Lab\\
	Shanghai Jiao Tong University\\
	kren, qinjr, zhenglei, zyyang, wnzhang, lqiu, yyu@apex.sjtu.edu.cn\\
}
\maketitle
\begin{abstract}
Survival analysis is a hotspot in statistical research for modeling time-to-event information with data censorship handling, which has been widely used in many applications such as clinical research, information system and other fields with survivorship bias. Many works have been proposed for survival analysis ranging from traditional statistic methods to machine learning models. However, the existing methodologies either utilize counting-based statistics on the segmented data, or have a pre-assumption on the event probability distribution w.r.t. time. Moreover, few works consider sequential patterns within the feature space. In this paper, we propose a Deep Recurrent Survival Analysis model which combines deep learning for conditional probability prediction at fine-grained level of the data, and survival analysis for tackling the censorship. By capturing the time dependency through modeling the conditional probability of the event for each sample, our method predicts the likelihood of the true event occurrence and estimates the survival rate over time, i.e., the probability of the \textit{non}-occurrence of the event, for the censored data. Meanwhile, without assuming any specific form of the event probability distribution, our model shows great advantages over the previous works on fitting various sophisticated data distributions. In the experiments on the three real-world tasks from different fields, our model significantly outperforms the state-of-the-art solutions under various metrics.
\end{abstract}

\section{Introduction}
Recent advances of modern technology makes redundant data collection available for \textit{time-to-event} information, which facilitates observing and tracking the event of interests.
However, due to different reasons, many events would lose tracking during observation period, which makes the data \textit{censored}.
We only know that the true time to the occurrence of the event is larger or smaller than, or within the observation time, which have been defined as survivorship bias categorized into \textit{right-censored}, \textit{left-censored} and \textit{internal-censored} respectively \cite{lee2003statistical}.
Survival analysis, a.k.a. time-to-event analysis \cite{lee2018deephit}, is a typical statistical methodology for modeling time-to-event data while handling censorship, which is a traditional research problem and has been studied over decades.

The goal of survival analysis is to estimate the time until occurrence of the particular event of interest, which can be regarded as a regression problem \cite{lee2003statistical,wu2015predicting}.
It can also be viewed as to predict the probability of the event occurring over the whole timeline \cite{wang2016functional,lee2018deephit}.
Specifically, given the information of the observing object, survival analysis would predict the probability of the event occurrence at each time point.

Nowadays, survival analysis has been widely used in real-world applications, such as clinical analysis in medicine research \cite{zhu2017wsisa,luck2017deep,katzman2018deepsurv} taking diseases as events and predicting survival time of patients; customer lifetime estimation in information systems \cite{jing2017neural,grob2018recurrent} which estimates the time until the next visit of users; market modeling in game theory fields \cite{wu2015predicting,wang2016functional} that predicts the event (i.e., winning) probability over the whole referral space.

Because of the essential applications in the real world, the researchers in both academic and industrial fields have devoted great efforts to studying survival analysis in recent decades.
Many works of survival analysis are from the view of traditional statistic methodology.
Among them, Kaplan-Meier estimator \cite{kaplan1958nonparametric} bases on non-parametric counting statistics and forecasts the survival rate at coarse-grained level where different observing objects may share the same forecasting result, which is not suitable in recent personalized applications.
Cox proportional hazard method \cite{cox1992regression} and its variants such as Lasso-Cox \cite{tibshirani1997lasso} assume specific stochastic process or base distribution with semi-parametric scaling coefficients for fine-tuning the final survival rate prediction.
Other parametric methods either make specific distributional assumptions, such as Exponential distribution \cite{lee2003statistical} and Weibull distribution \cite{ranganath2016deep}.
These methods pre-assume distributional forms for the survival rate function, which may not generalize very well in real-world situations.

Recently, deep learning, i.e., deep neural network, has been paid huge attention and introduced to survival analysis in many tasks \cite{ranganath2016deep,grob2018recurrent,lee2018deephit}.
However, in fact, many deep learning models for survival analysis \cite{katzman2018deepsurv,ranganath2016deep} actually utilize deep neural network as the enhanced feature extraction method \cite{lao2017deep,grob2018recurrent} and, worse still, rely on some assumptions of the base distributions for the survival rate prediction, which also suffers from the generalization problem.
Lately, \citet{lee2018deephit} proposed a deep learning method for modeling the event probability without assumptions of the probability distribution.
Nevertheless, they regard the event probability estimation as a pointwise prediction problem, and ignores the sequential patterns within neighboring time slices.
Moreover, the gradient signal is too sparse and has little effect on most of the prediction outputs of this model, which is not effective enough for modeling time-to-event data.

With the consideration of all the drawbacks within the existing literatures, in this paper we propose our Deep Recurrent Survival Analysis (DRSA) model for predicting the survival rate over time at fine-grained level, i.e., for each individual sample.
To the best of our knowledge, this is the first work utilizing auto-regressive model for capturing the sequential patterns of the feature over time in survival analysis.

Our model proposes a novel modeling view for time-to-event data, which aims at flexibly modeling the survival probability function rather than making any assumptions for the distribution form.
Specifically, DRSA creatively predicts the \textit{conditional} probability of the event at each time given that the event \textit{non}-occurred before, and combines them through probability chain rule for estimating both the probability density function and the cumulative distribution function of the event over time, eventually forecasts the survival rate at each time, which is more reasonable and mathematically efficient for survival analysis.
We train DRSA model by end-to-end optimization through maximum likelihood estimation, not only on the observed event among uncensored data, but also on the censored samples to reduce the survivorship bias.
Through these modeling methods, our DRSA model can capture the sequential patterns embedded in the feature space along the time, and output more effective distributions for each individual sample at fine-grained level.
The comprehensive experiments over three large-scale real-world datasets demonstrate that our model achieves significant improvements against state-of-the-art models under various metrics.

\section{Related Works}\label{sec:related-works}
\subsection{Learning over Censored Data}
The event occurrence information of some samples may be lost, due to some limitation of the observation period or losing tracks during the study procedure \cite{wang2017machine}, which is called data \textit{censorship}.
When dealing with time-to-event information, a more complex learning problem is to estimate the probability of the event occurrence at each time, especially for those samples without tracking logs after (or before) the observation time which is defined as \textit{right-censored} (or \textit{left-censored}) \cite{wang2017machine}.
Survival analysis is a typical statistical methodology for modeling time-to-event data while handling censorship.
There are two main streams of survival analysis.

The first view is based on traditional statistics scattering in three categories.
(i) Non-parametric methods including Kaplan-Meier estimator \cite{kaplan1958nonparametric} and Nelson-Aalen estimator \cite{andersen2012statistical} are solely based on counting statistics, which is too coarse-grained to perform personalized modeling.
(ii) Semi-parametric methods such as Cox proportional hazard model \cite{cox1992regression} and its variants Lasso-Cox \cite{tibshirani1997lasso} assumes some base distribution functions with the scaling coefficients for fine-tuning the final survival rate prediction.
(iii) Parametric models assume that the survival time or its logarithm result follows a particular theoretical distribution such as Exponential distribution \cite{lee2003statistical} and Weibull distribution \cite{ranganath2016deep}.
These methods either base on statistical counting information or pre-assume distributional forms for the survival rate function, which generalizes not very well in real-world situations.

The second school of survival analysis takes from machine learning perspective.
Survival random forest which was first proposed in \cite{gordon1985tree} derives from standard decision tree by modeling the censored data \cite{wang2016functional} while its idea is mainly based on counting-based statistics.
Other machine learning methodologies include Bayesian models \cite{ranganath2015survival,ranganath2016deep}, support vector machine \cite{khan2008support} and multi-task learning solutions \cite{Li2016mtlsa,alaa2017deep}.
Note that, deep learning models have emerged in recent years.
\citet{faraggi1995neural} first embedded neural network into Cox model to improve covariate relationship modeling.
From that, many works applied deep neural networks into well-studied statistical models to improve feature extraction and survival analysis through end-to-end learning, such as \cite{ranganath2016deep,luck2017deep,lao2017deep,katzman2018deepsurv,grob2018recurrent}.
Almost all the above models assume particular distribution forms which also suffers from the  generalization problem in practice.

\citet{biganzoli1998feed} utilized neural network directly to predict the survival rate for each sample and \citet{lisboa2003bayesian} extended it to a Bayesian network method.
In \cite{lee2018deephit} the authors proposed a feed forward deep model to directly predict the probability density values at each time point and sum them for estimating the survival rate.
However, in that paper, the gradient signal is quite sparse for the prediction outputs from the neural network.
Moreover, to our best knowledge, none of the related literatures considers the sequential patterns within the feature space over time.
We propose a recurrent neural network model predicting the \textit{conditional} probability of event at each time and estimate the survival rate through the probability chain rule, which captures the sequential dependency patterns between neighboring time slices and back-propagate the gradient more efficiently.

\subsection{Deep Learning and Recurrent Model}
Due to its adequate model capability and the support of big data, deep learning, a.k.a. deep neural network, has drawn great attention ranging from computer vision \cite{krizhevsky2012imagenet} and speech recognition \cite{graves2013speech} to natural language processing \cite{bahdanau2014neural} during the recent decades.
Among them, recurrent neural network (RNN) whose idea firstly emerged two decades ago and its variants like long short-term memory (LSTM) \cite{hochreiter1997long} employ memory structures to model the conditional probability which captures dynamic sequential patterns.
In this paper we borrow the idea of RNN and well design the modeling methodology for survival function regression.

\section{Deep Recurrent Survival Analysis}\label{sec:methodology}
In this section, we formulate the survival analysis problem and discuss the details of our proposed model.
We take the view of \textit{right-censorship} which is the most common scenario in survival analysis \cite{kaplan1958nonparametric,cox1992regression,wang2017machine,lee2018deephit}.

\subsection{Problem Definition}\label{sec:prob-def}
We define $z$ as the variable of the true occurrence time for the event of interest if it has been tracked.
We just simplify the occurrence of the event of interest as \textit{event}, and define the probabilistic density function (P.D.F.) of the true event time as $p(z)$, which means the probability that the event truely occurs at time $z$.

Now that we have the P.D.F. of the event time, we can derive the \textit{survival rate} at each time $t$ as the C.D.F. as
\begin{equation}\small
S(t) = \text{Pr}(z > t) = \int_t^{\infty} p(z) dz ~, \label{eq:survival-definition}
\end{equation}
which is the probability of the observing object surviving, i.e., event not occurring, until the observed time $t$.
Then the straightforward definition of the \textit{event rate}, 
i.e., the probability of event occurring before the observing time $t$, as that
\begin{equation}\small
W(t) = \text{Pr}(z \leq t) = 1 - S(t) = \int_0^t p(z) dz ~. \label{eq:win-prob-definition}
\end{equation}

The data of the survival analysis logs are represented as a set of $N$ triples $\{ \bx, z, t \}_{i=1}^N$, where $t > 0$ is the observation time for the given sample.
Here $z$ is left unknown (and we marked $z$ as null) for the censored samples without the observation of the true event time.
$\bx$ is the feature vector of the observation which encodes different information under various scenarios.

Our goal is to model the distribution of the true event time $p(z)$ over all the historical time-to-event logs with handling the censored data of which the true event time is unknown.
So the main problem of survival analysis is to estimate the probability distribution $p(z | \bx)$ of the event time with regard to the sample feature $\bx$, for each sample.
Formally speaking, the derived model is a ``mapping'' function $T$ which learns the patterns within the data and predicts the event time distribution over the time space as
\begin{equation}\label{eq:task-function}
p( z | \bx) = T(\bx) ~.
\vspace{-5pt}
\end{equation}

\subsection{Discrete Time Model}\label{sec:disc-time}
First of all, we present the definition of the \textit{conditional hazard rate} over continuous time as
\begin{equation}\label{eq:hazard}
h(t) = \lim_{\triangle t \rightarrow 0} \frac{\text{Pr}(t < z \leq t + \triangle t ~|~ z > t)}{\triangle t} ~,
\end{equation}
which models the instant occurrence probability of the event at time $t$ given that the event has not occurred before.
Note that the concept of \textit{hazard rate} has been commonly utilized in many survival analysis literatures \cite{cox1992regression,faraggi1995neural,luck2017deep}.

In the discrete context, a set of $L$ time slices $0 < t_1 < t_2 < \ldots < t_L$ is obtained which arises from the finite precision of time determinations.
Analogously we may also consider the grouping of continuous time as $l = 1, 2, \ldots, L$ and uniformly divide disjoint intervals $V_l = (t_{l-1}, t_{l}]$ where $t_0 = 0$ and $t_l$ is the last observation interval boundary for the given sample, i.e., the tracked observation time in the logs.
$V_L$ is the last time interval in the whole data space.
This setting is appropriately suited in our task and has been widely used in clinical research \cite{Li2016mtlsa,lee2018deephit}, information systems \cite{jing2017neural,grob2018recurrent} and other related fields \cite{wu2015predicting,wang2016functional}.

As such, our event rate function and survival rate function over discrete time space is
\begin{equation}\label{eq:win-lose-def}
\begin{aligned}
W(t_l) & = \text{Pr}(z \leq t_l) = \sum_{j \leq l} \text{Pr}(z \in V_j) ~, \\
S(t_l) & = \text{Pr}(z > t_l) = \sum_{j > l} \text{Pr}(z \in V_j) ~,
\end{aligned}
\vspace{-5pt}
\end{equation}
where the input to the two functions is the observed time $t_l$ from the log.
And the discrete event time probability function at the $l$-th time interval is
\begin{equation}\label{eq:discrete-pdf-def}\small
p_l = \text{Pr}(z \in V_l) = W(t_{l}) - W(t_{l-1}) = S(t_{l-1}) - S(t_{l}) ~.
\end{equation}

The discrete conditional hazard rate $h_l$, defined as the conditional probability as
\begin{equation}\label{eq:discrete-instant-win-func}\small
h_l = \text{Pr}(z \in V_l | z > t_{l-1}) = \frac{\text{Pr}(z \in V_l)}{\text{Pr}(z > t_{l-1})} = \frac{p_l}{S(t_{l-1})} ~,
\end{equation}
which approximates the continuous conditional hazard rate function $h(t_l)$ in Eq.~(\ref{eq:hazard}) as the intervals $V_l$ become infinitesimal.

\subsection{Deep Recurrent Model}\label{sec:drsa}
Till now, we have presented the discrete time model and discuss the death (i.e., event) and survival probability over the discrete time space.
We here propose our DRSA model based on recurrent neural network with the parameter $\btheta$, which captures the sequential patterns for conditional probability $h^i_l$ at every time interval $V_l$ for the $i^{\text{th}}$ sample.

\begin{figure}[t]
	\centering
	\includegraphics[width=1.0\columnwidth]{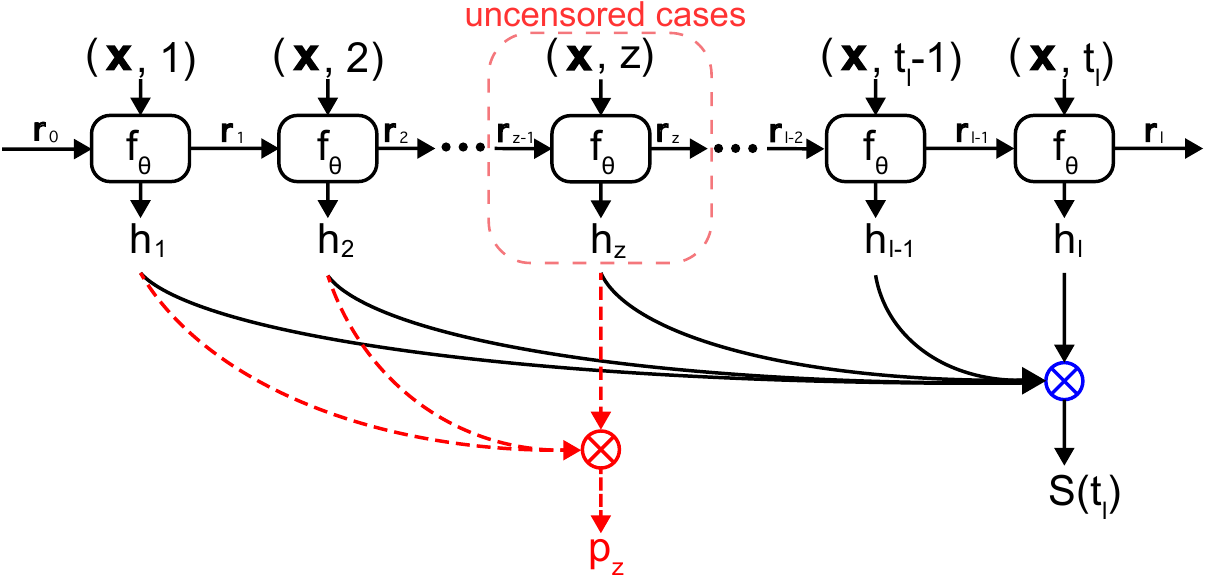}
	\caption{Detailed illustration of Deep Recurrent Survival Analysis (DRSA) model. Note that only the uncensored logs have the true event time and can calculate $p_z$ for the loss $L_z$. The calculation of $p_z$ and $S(t)$ have been derived in Eqs.~(\ref{eq:pdf-calc}) and (\ref{eq:discrete-survival-def}) respectively.}\label{fig:drsa}
	\vspace{-10pt}
\end{figure}

The detailed structure of DRSA network is illustrated in Figure~\ref{fig:drsa}.
At each time interval $V_l$, the $l$-th RNN cell predicts the instant hazard rate $h^i_l$ given the sample feature $\bx^i$ and the current time $t_l$ conditioned upon the previous events as
\begin{equation}\label{eq:discrete-hazard-function}
\begin{aligned}
h^i_l &= \text{Pr}(z \in V_l ~|~ z > t_{l-1}, ~ \bx^i; \btheta) \\
&= f_{\btheta}(\bx^i, t_l ~|~ \br_{l-1}) ~,
\end{aligned}
\end{equation}
where $f_{\btheta}$ is the RNN function taking $(\bx^i, t_l)$ as input and $h^i_l$ as output.
$\br_{l-1}$ is the hidden vector calculated from the RNN cell at the last time step which contains the information about the conditional.
It is quite natural for using the recurrent cell to model the conditional probability over time \cite{bahdanau2014neural}.
In our paper we implement the RNN function as a standard LSTM unit \cite{hochreiter1997long}, which has been widely used in sequence data modeling.
The details of the implementation of our RNN architecture can be referred in our supplemental materials and our reproductive code published in the experiment part.

From Eqs.~(\ref{eq:win-lose-def}), (\ref{eq:discrete-instant-win-func}) and (\ref{eq:discrete-hazard-function}), we can easily derive the survival rate function $S(t)$ through \textit{probability chain rule}, and the corresponding event rate function $W(t)$ at the time $t$ for the $i^{\text{th}}$ individual sample as
\begin{equation}\small
\begin{aligned}
S(t | \bx^i; \btheta) &= \text{Pr}(t < z | \bx^i; \btheta)  \\
&= \text{Pr}(z \not\in V_1, z \not\in V_2, \dots, z \not\in V_{l^i} | \bx^i; \btheta) \\
&= \text{Pr}(z \not\in V_1 |\bx^i; \btheta) \cdot \text{Pr}(z \not\in V_2 | z \not\in V_1, \bx^i; \btheta) \cdots \\
& ~~~~~~~~~~~~~\cdot \text{Pr}(z \not\in V_{l^i} | z \not\in V_1, \ldots, z \not\in V_{l^i-1}, \bx^i; \btheta) \\
& = \prod_{l: l \leq l^i} \left[ 1- \text{Pr}(z \in V_l ~|~ z > t_{l-1}  ,~ \bx^i; \btheta) \right] \\
&= \prod_{l: l \leq l^i} (1 - h^i_l) ~, \\ \label{eq:discrete-survival-def}
\end{aligned}
\end{equation}
\begin{equation}\small
\begin{aligned}
W(t | \bx^i; \btheta) &= \text{Pr}(t \geq z  | \bx^i; \btheta)= 1- \prod_{l: l \leq l^i} (1 - h^i_l)~, \label{eq:discrete-win-def}
\end{aligned}
\end{equation}
where $l^i$ is the time interval index for the $i^{\text{th}}$ sample at $t^i$.

Moreover, taking Eqs.~(\ref{eq:discrete-pdf-def}) and (\ref{eq:discrete-instant-win-func}) into consideration, the probability of time $z$ lying in the interval of $V_{l^i}$ for the $i^\text{th}$ sample is
\begin{equation}\label{eq:pdf-calc}
p^i_l = \text{Pr}(z \in V_{l^i} | \bx^i; \btheta) = h^i_{l^i} \prod_{l: l < l^i} (1 - h^i_l) ~.
\end{equation}

By means of probability chain rule, it connects all the outputs of the conditional hazard rate $h$ at each individual time to the final prediction, i.e., the probability $p(z)$ of the true event time $z$ and the survival rate $S(t)$ at each time $t$.
This feed-forward calculation guarantees that the gradient signal from the loss function can be transmitted through back-propagation more effectively comparing with \cite{lee2018deephit}, which will be discussed below.

\subsection{Loss Functions}
Since there is no ground truth of either the event time distribution or survival rate, here we maximize the log-likelihood over the empirical data distribution to learn our deep model.
Specifically, we take three objectives as our losses.

The first loss is to minimize the negative log-likelihood of the true event time $z = z^i$ over the \textit{uncensored} logs as
\begin{equation}\label{eq:market-price-objective}
\begin{aligned}
L_z &= - \log \prod_{(\bx^i, z^i) \in \mathbb{D}_{\text{uncensored}}} \text{Pr}(z \in V_{l^i} | \bx^i; \btheta) \\
&= - \log \prod_{(\bx^i, z^i) \in \mathbb{D}_{\text{uncensored}}} h^i_{l^i} \prod_{l: l < l^i} (1 - h^i_l) \\
&= - \sum_{(\bx^i, z^i) \in \mathbb{D}_{\text{uncensored}}} \left[ \log h^i_{l^i}  + \sum_{l: l < l^i} \log(1 - h^i_l)  \right] ~,
\end{aligned}
\end{equation}
where $l^i$ is the index of the interval of the true event time $z^i \in V_{l^i}$.

The second loss is to minimize the negative partial log-likelihood of the event rate over the \textit{uncensored} logs as
\begin{equation}\label{eq:winning-objective}
\begin{aligned}
L_{\text{uncensored}} &= - \log \prod_{(\bx^i, t^i) \in \mathbf{D}_{\text{uncensored}}} \text{Pr}(t^i \geq z | \bx^i; \btheta) \\
&= - \log \prod_{(\bx^i, t^i) \in \mathbb{D}_{\text{uncensored}}} W(t^i | \bx^i; \btheta) \\
&= - \sum_{(\bx^i, t^i) \in \mathbb{D}_{\text{uncensored}}} \log \Big[ 1- \prod_{l: l \leq l^i} (1 - h^i_l) \Big] ~.
\end{aligned}
\end{equation}
This loss adds more supervisions onto the predictions over the time range $(z^i,t^i)$ for the uncensored data than those \cite{katzman2018deepsurv,lee2018deephit,tibshirani1997lasso} merely supervise on the true event time $z^i$.

Though the censored logs do not contain any information about the true event time, we would only know that the true event time $z$ is greater than our logged observing time $t^i$ then.
Here we incorporate the partial log-likelihood embedded in the \textit{censored} logs as the third loss to correct the learning bias of our model as
\begin{equation}\label{eq:losing-objective}
\begin{aligned}
L_{\text{censored}} &= - \log \prod_{(\bx^i, t^i) \in \mathbb{D_{\text{censored}}}} \text{Pr}(z > t^i | \bx^i; \btheta) \\
&= - \log \prod_{(\bx^i, t^i) \in \mathbb{D_{\text{censored}}}} S(t^i | \bx^i; \btheta) \\
&= - \sum_{(\bx^i, t^i) \in \mathbb{D_{\text{censored}}}} \sum_{l: l \leq l^i} \log (1 - h^i_l) ~.
\end{aligned}
\end{equation}

\subsection{Model Realization}\label{sec:model-realization}
In this section, we unscramble some intrinsic properties of our deep survival model.

First of all, we analyze the model effectiveness of DRSA.
In \cite{lee2018deephit}, the proposed deep model directly predicts the event probability $p(t)$ and combines to estimate the survival rate as $S(t) = \sum_{t'\leq t}p(t')$ while ignoring the sequential patterns.
As a result, the gradient signal would only have effect on the prediction at time $t$ individually.
On the contrary, from Figure~\ref{fig:drsa}, we can see that our DRSA model is obviously more effective since the supervision would be directly back-propagated through the chain rule calculation to all the units with strict mathematical derivation, which guarantees to transmit the gradient more efficiently and effectively.
We also explicitly model the sequential patterns by conditional hazard rate prediction and we will illustrate the advantage of that in the experiments.

Then we take the view of censorship prediction of our methodology.
As is known that there is a censoring status as an indicator of survival at the given time, for each sample as
\begin{equation}\label{eq:lose-indicator}
\small
c^i(t^i)=
\begin{cases}
0, & \text{if}~~~~~~~~~~ t^i \geq z^i ~,
\cr 1, & \text{otherwise}~~ t^i < z^i ~.
\end{cases}
\end{equation}
In the tracking logs, each sample $(\bx^i, z^i, t^i)$ is uncensored where $c^i = 0$.
While for the censored logs losing tracking at the observation time, the true event time $z^i$ is unknown but the tracker only has the idea that $z^i > t^i$, thus $c^i = 1$.

Moreover, for the uncensored data, it is natural to ``push down'' the probability of survival $S(t^i)$. And for the censored data, it needs to ``pull up'' $S(t^i)$ since we ``observe event not occurred'' at time $t^i$.
However, using only $L_z$ to supervise the prediction of $p(z^i)$ at time $z^i$ in Eq.~(\ref{eq:market-price-objective}) is insufficient.
So that we incorporate the two partial likelihood losses $L_{\text{uncensored}}$ and $L_{\text{censored}}$ in Eqs.~(\ref{eq:winning-objective}) and (\ref{eq:losing-objective}).

Therefore, taking Eqs.~(\ref{eq:winning-objective}) and (\ref{eq:losing-objective}) altogether, we may find that the combination of $L_{\text{uncensored}}$ and $L_{\text{censored}}$ describes the classification of survival status at time $t^i$ of each sample as
\begin{align}
\tiny
& L_{\text{c}} = L_{\text{uncensored}} + L_{\text{censored}} \nonumber \\
&= - \log \prod_{(\bx^i, t^i) \in \mathbb{D}_{\text{full}}} \left[ S(t^i | \bx^i; \btheta) \right]^{c^i} \cdot \left[ 1 - S(t^i | \bx^i; \btheta) \right]^{1-c^i} \label{eq:classification-loss} \\
&= - \sum_{(\bx^i, t^i) \in \mathbb{D}_{\text{full}}} \left\{ c^i \cdot \log S(t^i | \bx^i; \btheta) \right. \nonumber \\
&~~~~~~~~~~~~~~~~~~~~~~~ + \left. (1-c^i) \log \left[ 1 - S(t^i | \bx^i; \btheta) \right] \right\} ~, \nonumber
\end{align}
which is the cross entropy loss for predicting the survival status at time $t^i$ given $\bx^i$ over all the data $\mathbb{D}_{\text{full}} = \mathbb{D}_{\text{uncensored}} \bigcup \mathbb{D}_{\text{censored}}$.

Combining all the objective functions and our goal is to minimize the negative log-likelihood over all the data samples including both uncensored and censored data as
\begin{equation}\label{eq:total-loss}
\arg \min_{\btheta}~~ \alpha L_z + (1-\alpha) L_{\text{c}} ~,
\end{equation}
where $\btheta$ is the model parameter in Eq.~(\ref{eq:discrete-hazard-function}) and the hyperparameter $\alpha$ controls the loss value balance between them.
Specifically, $\alpha$ controls the magnitudes of the two losses at the same level to stabilize the model training.

We also analyze the model efficiency in the supplemental material of this paper and the time complexity of model inference is the same as the traditional RNN model which has proven practical efficiency in the industrial applications \cite{zhang2014sequential}.

\section{Experiments}\label{sec:exps}
We evaluate our model with strong baselines in three real-world tasks.
Moreover, we have published the implementation code for reproductive experiments\footnote{Reproductive code link: https://github.com/rk2900/drsa.}.

\subsection{Tasks and Datasets}
We evaluate all the compared models in three real-world tasks.
We also published the processed full datasets\footnote{We have put sampled data in the published code. The three processed full datasets link: https://goo.gl/nUFND4.}.
\begin{description}
\item[CLINIC] is a dataset for tracking the patient clinic status \cite{knaus1995support}. Here the goal of survival analysis is to estimate the time till the event (death), and predict the probability of the event with waning effects of baseline physiologic variables over time.
\item[MUSIC] is a user lifetime analysis dataset \cite{jing2017neural} that contains roughly 1,000 users with entire listening history from 2004 to 2009 on last.fm, a famous online music service. Here the event is the user visit to the music service and the goal is to predict the time elapsed from the last visit of one user to her next visit.
\item[BIDDING] is a real-time bidding dataset in the computational advertising field \cite{ren2018bid,wang2016functional}.
In this scenario, the time is correspondent to the bid price of the bidder and the event is just winning of the auction. The feature contains the auction request information. Many researchers \cite{wu2015predicting,wang2016functional} utilized survival analysis for unbiased winning probability estimation of a single auction while handling the losing (censored) logs without knowing the true winning price.
\end{description}
The statistics of the three datasets are provided in Table~\ref{tab:stat}.
We split the CLINIC and MUSIC datasets to training and test sets with ratio of 4:1 and 6:1, respectively.
For feature engineering, all the datasets have been one-hot encoded for both categorical and numerical features.
The original BIDDING data have already been feature engineered and processed as training and test datasets.
Note that, the true time of the event of all the testing data have been preserved for the performance evaluation.
In these datasets, since all the time is integer value, we bucketize the discrete time interval as interval size $s_{\text{intv}}=1$
and the maximal time interval number $L$ is equal to the largest integer time in each dataset.
The discussion about various interval sizes has been included in the supplemental materials.

\begin{table}[t]
	\centering
	\caption{The statistics of the datasets. \#: number; AET: averaged event time. There are 9 independent subsets in BIDDING dataset so that we provide the overall statistics in this table and present the details in supplemental materials.}\label{tab:stat}
	\resizebox{\columnwidth}{!}{
		\begin{tabular}{|c|r|r|r|r|r|r|r|}
			\hline
		 	\multirow{2}*{Dataset} & \multirow{2}{*}{Total~\#~~} & Censored~ & {Censored} & AET~~~ & AET~~~~~~ & AET~~~~ & Feature \\
			& & Data \# ~~ & Rate~~~ & ($\mathbb{D}_{\text{full}}$)~~ & ($\mathbb{D}_{\text{uncensored}}$) & ($\mathbb{D}_{\text{censored}}$)~ & \# ~~~ \\
			\hline
			CLINIC & 6,036 & 797 & 0.1320 & 9.1141 & 5.3319 & 33.9762 & 14 \\
			MUSIC & 3,296,328 & 1,157,572 & 0.3511 & 122.1709 & 105.2404 & 153.4522 & 6 \\
			BIDDING & 19,495,974 & 14,848,243 & 0.7616 & 82.0744 & 25.0484 & 99.9244 & 12 \\
			\hline
		\end{tabular}
	}
\end{table}

\subsection{Evaluation Metrics}
The first evaluation metric is the \textit{time-dependent concordance index} (\textbf{C-index}), which is the most common evaluation used in survival analysis \cite{Li2016mtlsa,luck2017deep,lee2018deephit} and reflects a measure of how well a model predicts the ordering of sample event times.
That is, given the observing time $t$, two samples $d^1 = (\bx^1, z^1)$ with large event time $z^1 > t$ and $d^2 = (\bx^2, z^2)$ with small event time $z^2 \leq t$ should be ordered as $d^1 \prec d^2$ where $d^1$ is placed before $d^2$.
This evaluation is quite similar to the area under ROC curve (AUC) metric in the binary classification tasks \cite{wang2017machine}.
From the ranking view of event probability estimation at time $t$, C-index assesses the ordering performance among all the uncensored and censored pairs at $t$ among the test data.

We also use \textit{average negative log probability} (\textbf{ANLP}) to evaluate the regression performance among different forecasting models.
ANLP is to assess the likelihood of the co-occurrence of the test sample with the corresponding true event time, which is correspondent to the event time likelihood loss $L_z$ in Eq.~(\ref{eq:market-price-objective}).
Here we compute ANLP as
\begin{equation}
\bar{P} = - \frac{1}{|\mathbb{D}_{\text{test}}|} \sum_{(\bx^i, z^i) \in \mathbb{D}_{\text{test}}} \log p(z^i|\bx^i) ~,
\end{equation}
where $p(z|\bx)$ is the learned time-to-event probability function of each model.

Finally, we conduct the significance test to verify the statistical significance of the performance improvement of our model w.r.t. the baseline models.
Specifically, we deploy a MannWhitney U test \cite{mason2002areas} under C-index metric, and a t-test \cite{bhattacharya2002median} under ANLP metric.

\subsection{Compared Settings}
We compare our model with two traditional statistic methods and five machine learning methods including two deep learning models.

\begin{itemize}[leftmargin=2mm]
	\item \textbf{KM} is Kaplan-Meier estimator \cite{kaplan1958nonparametric} which is a statistic-based non-parametric method counting on the event probability at each time over the whole set of samples.
	\item \textbf{Lasso-Cox} is a semi-parametric method \cite{zhang2007adaptive} based on Cox proportional hazard model \cite{cox1992regression} with $l1$-regularization.
	\item \textbf{Gamma} is a parametric gamma distribution-based regression model \cite{zhu2017gamma}. The event time of each sample is modeled by a unique gamma distribution with respect to its features.
	\item \textbf{STM} is a survival random tree \cite{wang2016functional} model which splits the data into small segments using between-node heterogeneity and utilizes Kaplan-Meier estimator to derive the survival analysis for each segment.
	\item \textbf{MTLSA} is the recently proposed multi-task learning with survival analysis model \cite{Li2016mtlsa}. It transforms the original survival analysis problem into a series of binary classification subproblems, and uses a multi-task learning method to model the event probability at different time.
	\item \textbf{DeepSurv} is a Cox proportional hazard model with deep neural network \cite{katzman2018deepsurv} for feature extraction upon the sample covariates.
	\item \textbf{DeepHit} is a deep neural network model \cite{lee2018deephit} which predicts the probability $p(z)$ of event over the whole time space with the input $\bx$. This method achieved state-of-the-art performance in survival analysis problem.
	\item \textbf{DRSA} is our proposed model which has been described above. The implementation details can be referred to supplemental materials and the published code.

\end{itemize}

\subsection{Results and Analysis}
We present the evaluation results according to the category which the compared models belong to.
KM and Lasso-Cox model are statistic-based methods, while Gamma, STM and MTLSA are machine learning based models.
The rest models are deep neural network models with end-to-end learning paradigm.

\begin{table}[t]
	\centering
	\caption{Performance comparison on C-index (the \textit{higher}, the better) and ANLP (the \textit{lower}, the better). (* indicates p-value $<10^{-6}$ in significance test)}\label{tab:results}
	\resizebox{\columnwidth}{!}{
		\begin{tabular}{|c|c|c|c||c|c|c|}
			\hline
			\multirow{2}{*}{\textbf{Models}} & \multicolumn{3}{c||}{\textbf{C-index}} &  \multicolumn{3}{c|}{\textbf{ANLP}}\cr & CLINIC & MUSIC & BIDDING & CLINIC & MUSIC & BIDDING \cr
			\hline
			KM & 0.710 & 0.877 & 0.700 & 9.012 & 7.270 & 15.366 \cr
			Lasso-Cox & 0.752 & 0.868 & 0.834 & 5.307 & 28.983 & 38.620 \cr
			\hline\hline
			Gamma & 0.515 & 0.772 & 0.703 & 4.610 & 6.326 & 6.310 \cr
			STM & 0.520 & 0.875 & 0.807 & 3.780 & 5.707 & 5.148 \cr
			MTLSA & 0.643 & 0.509 & 0.513 & 17.759 & 25.121 & 9.668 \cr
			\hline\hline
			DeepSurv & 0.753 & 0.862 & 0.840 & 5.345 & 29.002 & 39.096 \cr
			DeepHit & 0.733 & 0.878 & 0.858 & 5.027 & 5.523 & 5.544 \cr
			DRSA & \bf 0.774\textsuperscript{*} & \bf 0.892\textsuperscript{*} & \bf 0.911\textsuperscript{*} & \bf 3.337\textsuperscript{*} & \bf 5.132\textsuperscript{*} & \bf 4.774\textsuperscript{*} \cr
			\hline
		\end{tabular}
	}
\end{table}

\subsubsection{Estimation for Event Rate over Time}
The left part of Table~\ref{tab:results} has illustrated the performance of the event rate estimation, i.e., C-index metric.
From the table, we may observe the following facts.
(i) Deep learning models illustrated relatively better C-index performance, which may be caused by the higher model capacity for feature extraction.
(ii) Not only within the deep learning models, but also over all the compared methods, our DRSA model achieved the best C-index scores with significant improvements on all the datasets, which proves the effectiveness of our model.
(iii) The traditional statistical models, i.e., KM and Lasso-Cox provided stable performance over the three datasets.
(iv) The models with pre-assumptions about the event time distribution, i.e., Gamma and DeepSurv, did not perform well because the strong assumptions lack generalization in real-world tasks.
(v) Within the deep models, no assumption about the latent distribution of time-to-event data makes DeepHit and DRSA flexibly model the data and perform better.

\subsubsection{Event Time Prediction}
ANLP is a metric to measure the regression performance on the true event time prediction, i.e., the forecasting of the likelihood of the true event time.

From the right part of Table~\ref{tab:results}, it also shows the similar findings to the C-index results discussed above, e.g., our DRSA model has the best performance among all the methods.
Moreover, STM segmented the data well so it achieved relatively better performance than other normal machine learning methods.
With effective sequential pattern mining over time, our DRSA model performed relatively better than other deep models. 
Note that DeepHit directly predicts the probability of time-to-event while our modeling method is based on hazard rate prediction and optimize through probability chain rule.
The results reflect the advantage of the sequential pattern mining with the novel modeling perspective of our model.

\begin{figure}[h]
	\centering
	\includegraphics[width=0.8\columnwidth]{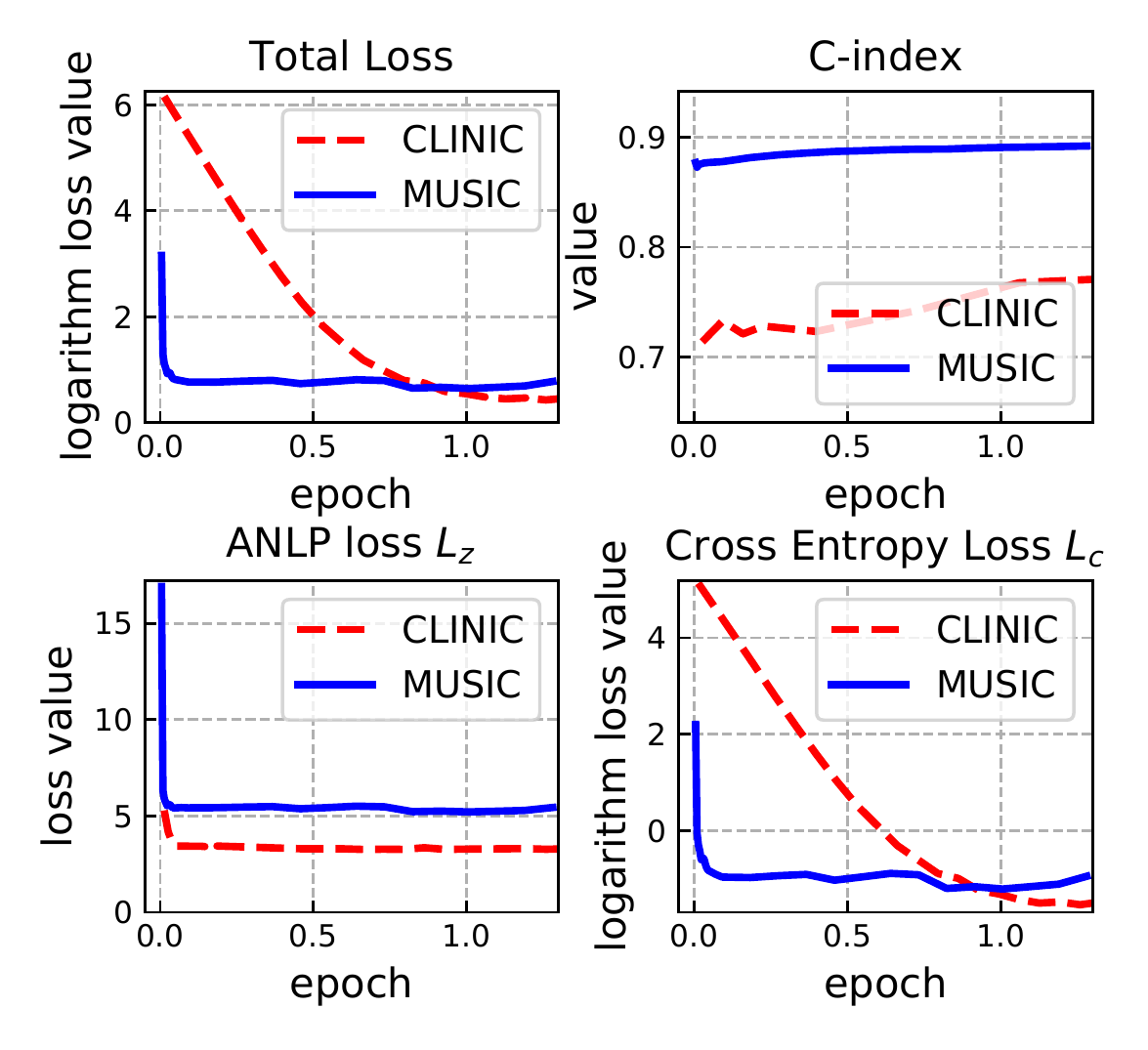}
	\caption{Learning curves. Here ``epoch'' means one iteration over the whole training data and $\alpha=0.25$ in Eq.~(\ref{eq:total-loss}). Learning curves over BIDDING dataset can be referred in our supplemental material.}\label{fig:learn-curve}
\end{figure}

\subsubsection{Model Convergence}
To illustrate the model training and convergence of DRSA model, we plot the learning curves and the C-index results on CLINIC and MUSIC datasets in Figure~\ref{fig:learn-curve}.
Recall that our model optimizes over two loss functions, i.e., the ANLP loss $L_z$ and the cross entropy loss $L_{\text{c}}$.
From the figure, we may find that DRSA converges quickly and the values of both loss function drop to stable convergence at about the first complete iteration over the whole training dataset.
Moreover, the two losses are alternatively optimizing and facilitate each other during the training, which proves the learning stability of our model.

\subsubsection{Model Forecasting Visualization}
\begin{figure}[t]
	\centering
	\includegraphics[width=1.0\columnwidth]{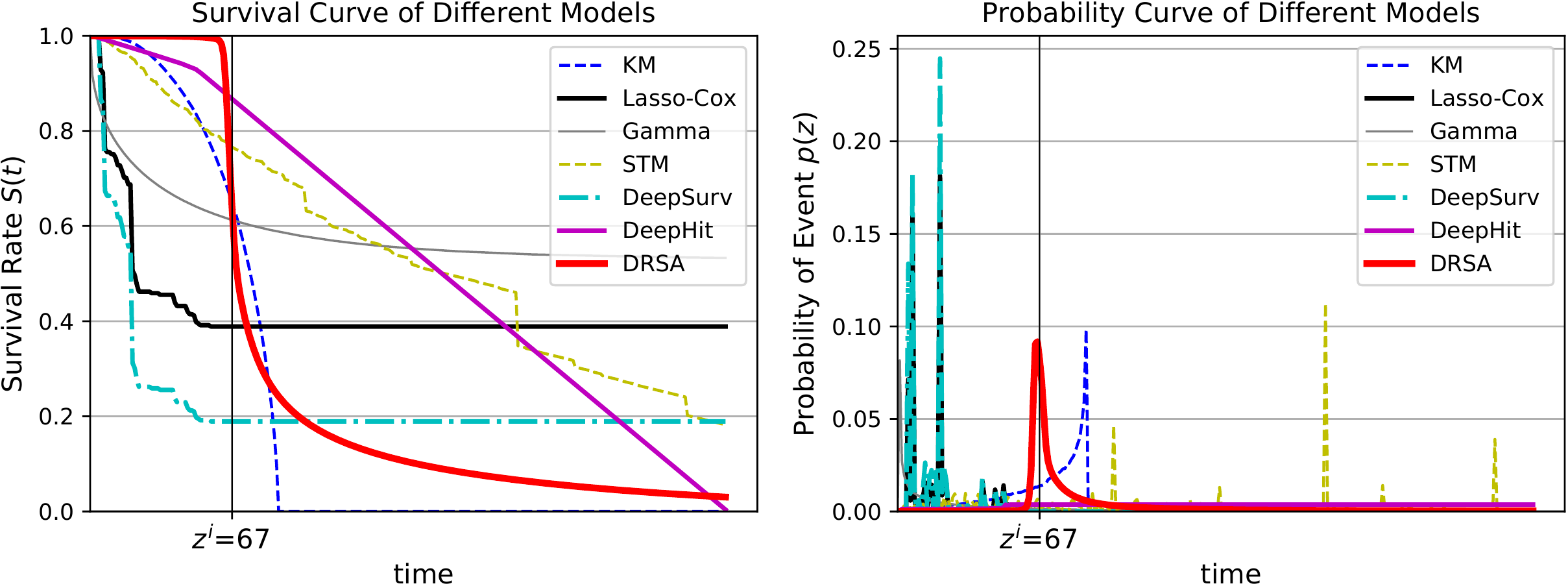}
	\caption{A comprehensive visualization of survival rate $S(t|\bx^i)$ estimation and event time probability $p(z|\bx^i)$ prediction over different models. The vertical dotted line is the true event time $z^i$ of this sample.}\label{fig:visualization}
\end{figure}
Figure~\ref{fig:visualization} illustrates the estimated survival rate curve $S(t|\bx^i)$ over time and the forecasted event time probability $p(z|\bx^i)$ for an arbitrarily selected test sample $(\bx^i, z^i, t^i)$.
Note that the KM model makes the same prediction for all the samples in the dataset, which is not personalized well.
Our DRSA model accurately placed the highest probability on the true event time $z^i$, which explains the result of ANLP metric where DRSA achieved the best ANLP scores.
Since DeepHit directly predicted the probability of the event time $p(z^i)$ without any considerations of the previous conditional.
And it has no supervision onto the predictions in the time range $(z^i, t^i)$ which makes the gradient signal too sparse only onto the true event time $z^i$.
As a result, DeepHit did not place the probability well over the whole time space.

\subsubsection{Ablation Study on the Losses}
\begin{table}[t]
	\centering
	\caption{Ablation study on the losses.}\label{tab:ab-results}
	\resizebox{\columnwidth}{!}{
		\begin{tabular}{|c|c|c|c||c|c|c|}
			\hline
			\multirow{2}{*}{\textbf{Models}} & \multicolumn{3}{c||}{\textbf{C-index}} &  \multicolumn{3}{c|}{\textbf{ANLP}}\cr & CLINIC & MUSIC & BIDDING & CLINIC & MUSIC & BIDDING \cr
			\hline
			DeepHit & 0.733 & 0.878 & 0.858 & 5.027 & 5.523 & 5.544 \cr
			DRSA$_{\text{unc}}$ & 0.765 & 0.881 & 0.823 & 3.441 & 5.412 & 12.255 \cr
			DRSA$_{\text{cen}}$ & 0.760 & 0.882 & 0.900 & \textbf{3.136} & 5.459 & 4.990 \cr
			DRSA & \bf 0.774 & \bf 0.892 & \bf 0.911 & 3.337 & \bf 5.132 & \bf 4.774 \cr
			\hline
		\end{tabular}
	}
\end{table}
In this ablation study, we compare the model performance on the three losses. DRSA$_{\text{unc}}$ optimizes under $(L_z + L_{\text{uncensored}})$ over only the uncensored data, and DRSA$_{\text{cen}}$ optimizes under $(L_z + L_{\text{censored}})$ without the loss $L_{\text{uncensored}}$.
Note that our full model DRSA optimizes under all the three losses $(L_z + L_{\text{uncensored}} + L_{\text{censored}})$ as stated in Eq.~(\ref{eq:total-loss}).
From Table~\ref{tab:ab-results}, we may find that both two partial likelihood losses $L_{\text{uncensored}}$ and $L_{\text{censored}}$ contribute to the final prediction.
Moreover, our DRSA over all the three losses achieved the best performance, which reflects the effectiveness of our classification loss $L_c = L_{\text{uncensored}} + L_{\text{censored}}$ as that in Eq.~(\ref{eq:classification-loss}), which optimizes the C-index metric directly.

\section{Conclusion}\label{sec:conclusion}
In this paper, we comprehensively surveyed the survival analysis works from the modeling view and discussed the pros and cons of them.
To make flexibly modeling over time, we proposed a deep recurrent neural network with novel modeling view for conditional hazard rate prediction.
And probability chain rule connects the predicted hazard rate at each time, for the event time probability forecasting and survival rate estimation.
The experiments on three large-scale datasets in three real-world tasks from different fields illustrated the significant advantages of our model against the strong baselines including state-of-the-art model.

For the future work, it is natural to apply our model for competing risks prediction \cite{alaa2017deep,lee2018deephit} with shared feature embedding at the base architecture and multi-task learning for loss function.

\section{Acknowledgments}
The corresponding authors Weinan Zhang and Yong Yu thank the support of National Natural Science Foundation of China (61632017, 61702327, 61772333), Shanghai Sailing Program (17YF1428200).

\bibliography{deep-survival}
\bibliographystyle{aaai}

\end{document}